# Recognition-Oriented Low-Light Image Enhancement based on Global and Pixelwise Optimization

Seitaro Ono[1], Yuka Ogino[2], Takahiro Toizumi[2], Atsushi Ito[2] and Masato Tsukada[1]
[1] *University of Tsukuba, Ibaraki, Japan*
[2] *NEC Corporation, Kanagawa, Japan*
*seitaro.ono@image.iit.tsukuba.ac.jp, tsukada@iit.tsukuba.ac.jp, {yogino, t-toizumi_ct, ito-atsushi}@nec.com,*

Abstract: In this paper, we propose a novel low-light image enhancement method aimed at improving the performance of recognition models. Despite recent advances in deep learning, the recognition of images under low-light conditions remains a challenge. Although existing low-light image enhancement methods have been developed to improve image visibility for human vision, they do not specifically focus on enhancing recognition model performance. Our proposed low-light image enhancement method consists of two key modules: the Global Enhance Module, which adjusts the overall brightness and color balance of the input image, and the Pixelwise Adjustment Module, which refines image features at the pixel level. These modules are trained to enhance input images to improve downstream recognition model performance effectively. Notably, the proposed method can be applied as a frontend filter to improve low-light recognition performance without requiring retraining of downstream recognition models. Experimental results demonstrate that our method improves the performance of pretrained recognition models under low-light conditions and its effectiveness.

## 1 INTRODUCTION

In recent years, deep learning-based image recognition models have achieved significant advancements (Wang et al., 2020; Muhammad et al., 2022; Zheng et al., 2023), demonstrating progressively improved performance. However, these models highly rely on large datasets (Geiger et al., 2012; Andriluka et al., 2014; Lin et al., 2014; Güler et al., 2018) consisting of high-quality images captured under ideal good lighting conditions, presenting substantial challenges when applied to low-light environments (Tian et al., 2023; Ono et al., 2024). In low-light settings, factors such as low exposure and reduced contrast severely impair image visibility. To improve the visibility of low-light images, deep learning-based low-light image enhancement methods have been developed (Wang et al., 2020; Li et al., 2021; Liu et al., 2021; Tian et al., 2023). However, these methods are primarily designed for human vision, without consideration for the performance of recognition models. Consequently, conventional enhancement methods may cause excessive smoothing or amplifying noise in output images, potentially degrading recognition model performance (Ono et al., 2024; Ogino et al., 2024).

To address the reduction in recognition performance under low-light conditions, we propose a novel low-light image enhancement method that converts input images into more recognizable images for downstream recognition models. The proposed enhancement method comprises two modules: the Global Enhancement Module (GEM), which adjusts the overall brightness of input images, and the Pixelwise Adjustment Module (PAM), which refines image features on a pixel level. GEM adjusts the brightness of the entire input image, producing a globally enhanced image, $I_{global}$. PAM, on the other hand, estimates an optimal pixel-wise correction map, $f_{local}$, that records adjustment values for each pixel in $I_{global}$. By taking the linear combination of $I_{global}$ and $f_{local}$, we generate the output image $I_{out}$.

Our enhancement model is trained to minimize a model specific loss function for a downstream recognition model by optimizing the output image $I_{out}$ to make it more suitable for the downstream recognition model. By optimizing our low-light enhancement method to reduce the model specific loss for the downstream recognition model, we enhance the input image in a way that facilitates improved recognition performance.

To evaluate the effectiveness of our proposed method under low-light conditions and its generalizability across different recognition tasks, we conducted experiments on two tasks: single-person pose estimation and semantic segmentation. The experimental results demonstrate that our proposed method effectively enhances the performance of pretrained recognition models under low-light conditions and improves their performance across different recognition tasks, validating its effectiveness and generalizability.

## 2 RELATED WORKS

Image recognition under low-light conditions is a critical challenge in computer vision (Liang et al., 2021; Wang et al., 2022), and various methods have been proposed to improve image visibility in such environments. In recent years, various low-light image enhancement techniques based on convolutional neural networks have emerged (Wang et al., 2020; Li et al., 2021; Tian et al., 2023). Zero-DCE (Guo et al., 2020) formulates low-light image enhancement as an image-specific tone curve estimation task in a deep network. By designing a non-reference loss function, Zero-DCE can be effectively trained without paired datasets and has been shown to perform well across diverse lighting conditions.

LLFLow (Wang et al., 2022) is a supervised low-light image enhancement framework based on normalizing flow, designed to model the distribution of well-exposed images accurately. By utilizing this approach, LLFLow achieves enhanced structural detail preservation across varied contexts, leading to superior restoration quality in low-light conditions.

Although these conventional methods are effective for enhancing low-light images, many of these data-driven methods rely heavily on large-paired data sets of low-light and bright images (Cai et al., 2018; Chen et al., 2018; Wei et al., 2018). Not only is the collection of such data sets very costly, but this dependence also limits their practicality by limiting the situations in which they can be applied.

Furthermore, these methods are mainly designed to improve image visibility for human vision and often do not consider their impact on downstream recognition models. Consequently, applying these techniques to low-light images can unintentionally discard features vital for the downstream recognition model during the enhancement process (Ono et al., 2024).

The Image-Adaptive Learnable Module (IALM) (Ono et al., 2024) is a novel low-light image enhancement method explicitly designed to improve downstream recognition model performance rather than human perceptibility. Unlike conventional enhancement methods, which minimize a distance function between restored images and ground-truth images, IALM maximizes recognition performance by optimizing image correction applied to input images so as to minimize the loss function specific to the downstream recognition model. This approach ensures that the enhancement process adapts input images to improve recognition model performance, effectively enhancing features essential for recognition.

IALM consists of three image processing modules: the Exposure Module, which adaptively adjusts the exposure of an input image to suit the downstream recognition model; the Gamma Module, which adaptively adjusts the image contrast; and the Smoothing Module, which performs noise reduction. Each of these modules includes a parameter predictor that estimates the image processing parameters needed for each adjustment. Although IALM utilizes a relatively simple image processing module, it has been shown to improve the performance of downstream recognition models compared to conventional low-light image enhancement methods (Ono et al., 2024). However, IALM only supports global image adjustment, which imposes a limitation in that it cannot perform detailed pixel-level correction for low-light images. Consequently, to further enhance downstream recognition model performance, it is necessary to develop low-light image enhancement methods tailored to recognition models, equipped with more fine-grained image processing capabilities.

## 3 PROPOSED METHOD

IALM performs exposure correction, gamma adjustment, and noise reduction on input low-light images in accordance with the characteristics of a downstream recognition model. By uniformly adjusting the entire image, IALM enhances recognition model performance. We think that applying pixelwise feature adjustment may further improve the performance of the recognition model. In this study, we propose a low-light image enhancement method designed to improve recognition performance better, comprising two modules: the Global Enhancement Module (GEM), which globally adjusts the brightness and the color balance of an input image, and the Pixelwise Adjustment Module (PAM), which performs pixelwise corrections to enhance image features. An overview of our proposed framework is shown in Figure 1.

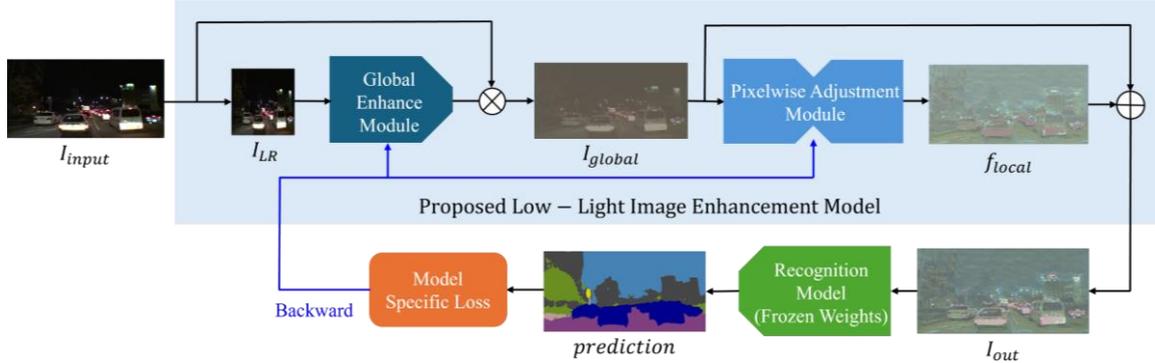

Figure 1: The overall training pipeline for the proposed Low-Light Image Enhancement method. Input low-light image is fed to GEM for global correction of exposure and color balance. The corrected image is then fed into the PAM, which corrects the image at the pixel level, yielding the final output image. The output image is then input to the downstream recognition model (Frozen Weights), where model specific losses are calculated. The weights of the proposed low-light image enhancement model are updated to minimize model-specific losses. This process corrects the images to improve the performance of the downstream recognition model.

An input image $I_{input}$ is first downsampled using bilinear interpolation, yielding a low-resolution image $I_{LR}$. $I_{LR}$ is then fed into GEM, which determines the optimal correction parameters to adjust the $I_{input}$ linearly. While downsampling reduces high-frequency components of images, it allows GEM to focus on capturing global information, such as exposure. By applying GEM's predicted correction parameters to the $I_{input}$, we obtain a globally enhanced image $I_{global}$. $I_{global}$ is then passed to PAM, which creates a pixelwise adjustment map, $f_{local}$, that refines image features beneficial to a downstream recognition model. The correction map $f_{local}$ is added to $I_{global}$, resulting in the output image $I_{out}$. The output image, $I_{out}$, is fed into the recognition model, where a model-specific loss is calculated based on the difference between the model's predictions and the ground truth. After calculating this loss, the low-light image enhancement model is trained to minimize it, thereby improving image features in a way that enhances recognition performance.

Since our method focuses on optimizing the image enhancement method as a pre-processing step for the recognition model, no gradient updates are applied to the recognition model during training. This training strategy leverages the pretrained knowledge of the recognition model without requiring its retraining. It effectively highlights image features beneficial to the recognition model and boosts recognition performance. Our low-light enhancement method is implemented with a compact convolutional neural network architecture containing only 577k parameters. Compared to conventional low-light enhancement methods, which often require millions of parameters and multiple convolutional layers, our method is lightweight, highly practical, and suitable for a wide range of real-world applications.

### 3.1 Global Enhance Module

Global Enhancement Module (GEM) adaptively adjusts the global brightness and color balance of an input image. GEM is a lightweight convolutional neural network composed of six convolutional layers. Given that processing high-resolution images with convolutional neural networks requires substantial computational resources, GEM is fed a low-resolution image, $I_{LR}$, created by downsampling the input image $I_{input}$ to a 32×32 resolution using bilinear interpolation to reduce the resource demands.

Although the downsampling process removes high-frequency features from $I_{input}$, the primary goal of GEM is to understand global information such as exposure and calculate adaptive exposure correction parameters for $I_{input}$; thus, a low-resolution input is sufficient and appropriate for this purpose.

The low-resolution image $I_{LR}$ is fed into GEM, which then predicts three correction coefficients $(a_R, a_G, a_B)$ by a convolutional neural network for adjusting the exposure and the color balance of $I_{input}$. For each pixel $i$ in $I_{input}$, with initial pixel values $(r_i, g_i, b_i)$ and adjusted pixel values $(r'_i, g'_i, b'_i)$, the following operation is performed.

$$(r'_i, g'_i, b'_i) = (a_R r_i, a_G g_i, a_B b_i) \quad (1)$$

This correction yields the exposure-enhanced and color-balanced image, $I_{global}$.

## 3.2 Pixelwise Adjustment Module

Pixelwise Adjustment Module (PAM) is designed to predict a pixelwise adjustment map that enhances beneficial image features to improve the performance of downstream recognition models.

To obtain features beneficial for downstream recognition models at the pixel level, PAM utilizes a tiny fully convolutional neural network based on the UNet (Ronneberger et al, 2015) architecture. This fully convolutional network has a symmetric encoder and decoder structure.

The encoder part consists of four convolutional blocks. Each convolutional block of the encoder consists of two convolutional layers, a batch normalization layer, and a ReLU activation function to extract important information while reducing the resolution of the feature map.

The decoder section consists of four transposed convolutional blocks. Each transposed convolution block of the decoder section consists of a transposed convolution layer, a convolution layer, a batch-normalization layer, and a ReLU activation function.

The UNet architecture allows fine pixel information to be refined while gradually increasing the resolution of the feature map obtained from the encoder at the decoder. Each transposed convolutional block combines features from the corresponding encoder layer via skip connections. Finally, we obtain a pixelwise adjustment map of the same size as the input image size.

In the image processing workflow of PAM, the exposure-corrected and color-balanced image $I_{global}$, produced by GEM is first fed into PAM. The PAM then predicts a pixelwise adjustment map of the same resolution as $I_{global}$. The predicted pixelwise adjustment map, $f_{local}$, is added to $I_{global}$ to generate the output image $I_{out}$, as described by the following operation:

$$I_{out} = I_{global} + f_{local} \quad (2)$$

This process produces an output image optimized to improve performance in downstream recognition models.

## 4 EXPERIMENTS

We demonstrated the effectiveness of our proposed method under low-light conditions. We quantitatively evaluated the performance of the proposed method in low-light conditions. We conducted experiments across different tasks to demonstrate the generalizability of the proposed method. We also compared the enhanced images with those processed by conventional methods for a qualitative evaluation. We also conducted an ablation study to evaluate the effectiveness of each of the two proposed modules.

## 4.1 Implementation Details

We evaluated the effectiveness and generalizability of our proposed method using two different recognition tasks: single person pose estimation and semantic segmentation.

Single person pose estimation involves predicting the coordinates of keypoints, such as the head, shoulders, and elbows, for a single person present in the input image (Zheng et al., 2023). On the other hand, semantic segmentation aims to predict pixel-wise classifications for various class categories present in the image (Garcia-Garcia et al., 2018).

For the single person pose estimation task, we used the pose estimation model proposed by Lee et al. (Lee et al., 2023) as our recognition model. This model is pre-trained on the ExLPose dataset (Lee et al., 2023), specifically designed for pose estimation in low-light conditions, enabling it to perform pose estimation in both dark and well-lit environments. We evaluated the performance of Lee et al.'s pose estimation model both with and without applying our proposed LLIE method.

In the semantic segmentation task, we utilized DeepLabV3+ (Chen et al., 2018) as a recognition model. DeepLabV3+ has been pre-trained on the Cityscapes dataset (Cordts et al., 2016), which consists of semantic segmentation data collected from daytime urban street scenes. Unlike the model by Lee et al., DeepLabV3+ is not designed for estimation in low-light conditions. To evaluate the effectiveness of our proposed method under low-light conditions, we used the NightCity dataset (Tan et al., 2021), which comprises nighttime city driving scenes, as test data in the semantic segmentation task. Similar to the single person pose estimation task, we evaluated our method's effectiveness by comparing the performance of the pre-trained DeepLabV3+ on low-light test images from NightCity with and without our method.

During the training of our proposed method, we froze the weights of the downstream recognition model in both recognition tasks and focused solely on training the low-light image enhancement method. To evaluate the effectiveness of our method against conventional methods, we compared it with Zero-DCE, LLFLow, and IALM. For the single person pose estimation task, we trained these methods using images from the ExLPose dataset. In the Semantic Segmentation task, we trained each low-light image enhancement method on the training data from the NightCity dataset. Since the NightCity dataset does not contain pairs of bright images corresponding to low-light images, LLFLow, which is a supervised method, cannot be trained. So, we used LLFLow trained on the LOL dataset (Wei et al., 2018) in this experiment. The LOL dataset is a widely used benchmark dataset for low-light image enhancement, containing 500 pairs of low-light and bright images primarily captured indoors.

During the training of our proposed method, we adopted the Adam Optimizer. We set the learning rate to $5 \times 10^{-4}$ and the batch size to 8. We conducted all experiments using PyTorch, with training performed on an RTX 3060.

### 4.2 Datasets

#### 4.2.1 ExLPose

The ExLPose dataset is designed for human pose estimation under low-light conditions. It provides pairs of bright and low-light images. The dataset includes 2,065 training pairs and 491 test pairs, along with annotations for bounding boxes and keypoint coordinates of the individuals in the images. The test data is categorized into four subsets based on mean pixel intensity: LL-N (Normal), LL-H (Hard), LL-E (Extreme), and LL-A (All). The mean pixel intensity for each subset is 3.2, 1.4, 0.9, and 2.0, respectively. In this study, we utilized the training data for pre-training the recognition model and training all low-light image enhancement methods. During the evaluation experiments, we conducted assessments on all subsets of the ExLPose's test data.

#### 4.2.2 Cityscapes

The Cityscapes dataset consists of images of daytime urban street scenes for semantic segmentation. It serves as a benchmark dataset for segmentation tasks. The dataset includes 2,975 training images, 500 validation images, and 1,525 test images, along with pixel-wise annotations for 19 categories. Each image has a resolution of $2048 \times 1024$ pixels. In this study, we utilized only the training images for pre-training DeepLabV3+.

#### 4.2.3 NightCity

The NightCity dataset consists of images of nighttime urban street scenes, designed for semantic segmentation tasks. This dataset provides 2,998 training images with a size of $1024 \times 512$ pixels and 1,299 test images. It also offers pixelwise annotations for the same 19 categories as the Cityscapes dataset. In this study, we utilized this dataset for both training and testing all low-light image enhancement methods.

### 4.3 Evaluation Protocol

For the single person pose estimation task, we adopted the Average Precision (AP) score based on object keypoint similarity (OKS) as our evaluation metric. We reported the mean AP over OKS thresholds ranging from 0.5 to 0.95, with a step size of 0.05. The AP value increases with higher accuracy in keypoint estimation.

For the semantic segmentation task, we adopted the mean of category-wise intersection-over-union (mIoU) as the evaluation metric. The mIoU value increases with higher accuracy in segmentation, indicating better recognition performance.

### 4.4 Experimental Results

First, we present the experimental results for the single person pose estimation task in Table 1. The results show that applying conventional low-light image enhancement methods, such as Zero-DCE and LLFLow, which do not account for the characteristics of downstream recognition models, led to decreased recognition performance compared to when no enhancement was applied.

In contrast, IALM, which was trained to maximize the prediction performance of the downstream recognition model, successfully improved the performance across all subsets of the test data. Due to the limitations of IALM's correction capabilities, the improvement in performance was marginal.

On the other hand, our proposed method achieved even higher performance on all subsets of the ExLPose's test data compared to conventional methods, confirming its effectiveness.

We also present in Table 2 the number of parameters for each method as well as the latency required to process a single input image on an RTX3060 GPU. The input images in the single person pose estimation task were $256 \times 192$-pixel RGB images.

According to Table 2, the proposed method demonstrates significantly faster performance than other competing methods. Our method also has relatively fewer parameters than conventional methods that require many convolutional layers, such as LLFLow. Therefore, it can effectively improve the performance of existing trained recognition models in low-light conditions at a low computational cost.

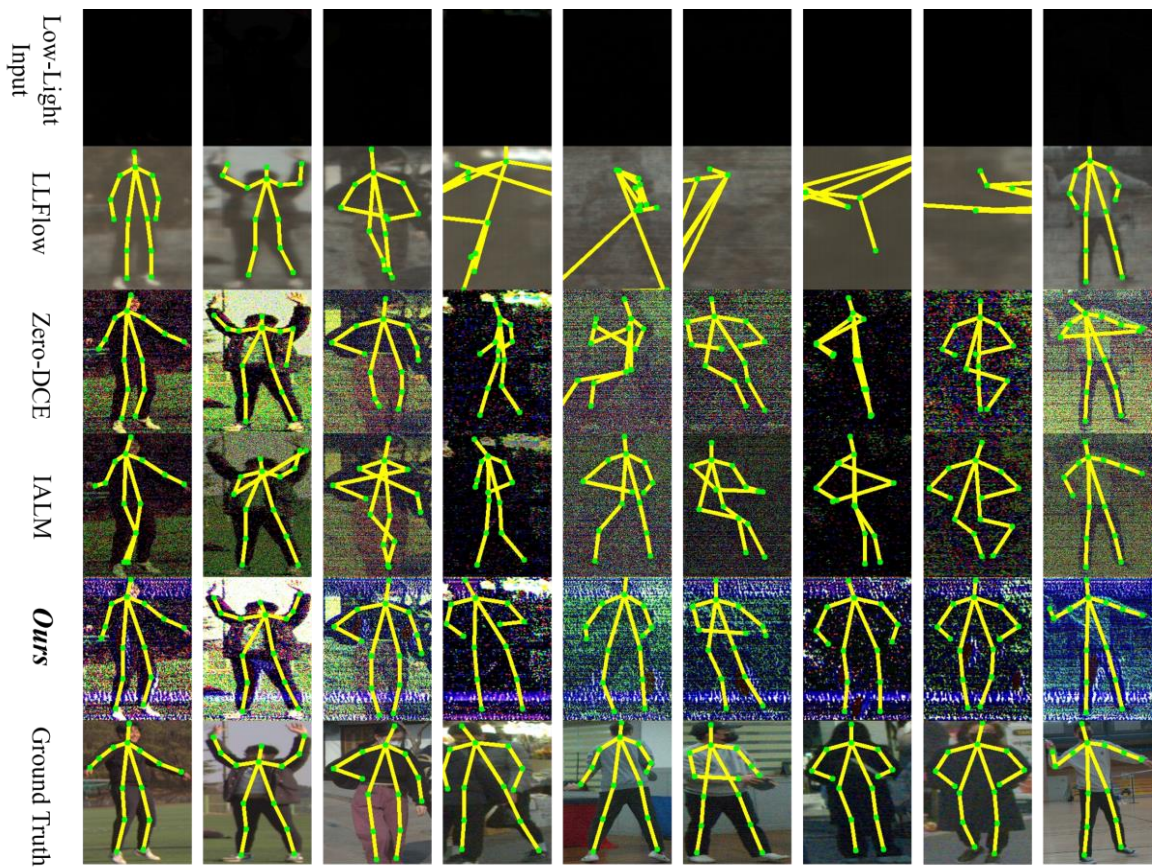

Figure 2: Qualitative evaluation results of the single person pose estimation task. Conventional low-light image enhancement methods can increase image visibility, but do not necessarily improve the performance of the downstream recognition model. The output images obtained with our method has purple noise, which is not high quality for human vision. However, it allows for highly accurate pose estimation.

Table 1: Quantitative evaluation results of the single person pose estimation task.

| | AP@0.5-0.95 ↑ | | | |
|---|---|---|---|---|
| Model | LL-N | LL-H | LL-E | LL-A |
| Lee et al. | 42.1 | 33.8 | 18.0 | 32.4 |
| LLFLow and Lee et al. | 9.3 | 4.6 | 1.2 | 5.4 |
| Zero-DCE and Lee et al. | 34.7 | 28.6 | 16.7 | 27.5 |
| IALM and Lee et al. | 42.6 | 34.1 | 20.0 | 33.2 |
| *Ours and Lee et al.* | **43.6** | **34.4** | **21.0** | **34.1** |

Table 2: Comparison of the processing speed of each method in the single person pose estimation task. Input images are RGB images of size $256 \times 192$-pixel in this task.

| Model | Params [M] | Latency [ms] |
|---|---|---|
| Lee et al. | 27.4 | 13.8 |
| LLFLow and Lee et al. | 66.3 | 339.8 |
| Zero-DCE and Lee et al. | 27.5 | 16.5 |
| IALM and Lee et al. | 27.9 | 18.7 |
| *Ours and Lee et al.* | 28.0 | **16.4** |

Figure 2 illustrates the qualitative evaluation results for the single person pose estimation task. The experimental results indicated that LLFlow demonstrates excessive smoothing in the output images, while Zero-DCE introduces additional noise, which may contribute to recognition errors. IALM successfully adapted the exposure to align with the characteristics of the downstream recognition model and managed to suppress noise. However, some cases of recognition failure still occurred. Our proposed method generated purple artifacts at the top and bottom of the output images, which may not be visually pleasing for human vision. Nonetheless, it achieved higher recognition performance compared to conventional methods.

The experimental results for the semantic segmentation task are shown in Table 3. Similar to the previous task, conventional low-light image enhancement methods Zero-DCE and LLFLow, which do not take into account the characteristics of the downstream recognition model, led to a reduction in performance. In contrast, IALM achieved an improvement in performance, although the improvement was marginal.

Our proposed method successfully improved the performance of DeepLabV3+, which was trained solely on images captured during the day, achieving approximately a 1.87-fold improvement under low-light conditions. We also show in Table 4 the number of parameters for each method and the latency required to process one input image on the RTX3060 GPU for the semantic segmentation task. The input images for the semantic segmentation task were 1024 × 512-pixel RGB images. As shown in Table 4, the proposed method exhibited superior speed compared to other competitive methods, even when handling large 1024 × 512-pixel input images.

Figure 3 presents the qualitative evaluation results for the semantic segmentation task. The qualitative assessment indicated that conventional methods Zero-DCE and LLFLow, which aim to improve image quality for human vision, produced visually appealing enhancement results but led to incorrect recognition outcomes. IALM also exhibited instances of erroneous recognition. While the images enhanced by our proposed method did not achieve visually appealing results, they demonstrated an improvement in recognition performance.

These experiments collectively validated the effectiveness of our proposed method and its generalizability across different tasks.

### 4.5 Ablation Study

GEM performs exposure correction and color balance adjustment by adaptively multiplying a constant value for each channel of the input image. However, this simple image processing can be implicitly modeled by multiple convolution operations by the PAM at a later stage. In that case, it may not be necessary to include GEM before PAM. We considered this possibility and conducted an ablation study to validate the effectiveness of each of the two proposed modules.

In this experiment, we adopted the semantic segmentation task as a recognition task and DeepLabV3+ as a downstream recognition model and reported the mIoU when the proposed method was trained excluding GEM and PAM, respectively.

Table 3: Quantitative evaluation results of the semantic segmentation task.

| Model | mIoU ↑ |
| --- | --- |
| DeepLabV3+ | 18.4 |
| LLFLow and DeepLabV3+ | 15.3 |
| Zero-DCE and DeepLabV3+ | 16.7 |
| IALM and DeepLabV3+ | 18.7 |
| *Ours and DeepLabV3+* | *34.4* |

Table 4: Comparison of the processing speed of each method in the semantic segmentation task. Input images are RGB images of size 1024 × 512-pixel in this task.

| Model | Params [M] | Latency [ms] |
| --- | --- | --- |
| DeepLabV3+ | 5.2 | 18.1 |
| LLFLow and DeepLabV3+ | 44.1 | 1,027.9 |
| Zero-DCE and DeepLabV3+ | 5.3 | 38.2 |
| IALM and DeepLabV3+ | 5.7 | 32.1 |
| *Ours and DeepLabV3+* | 5.8 | *26.3* |

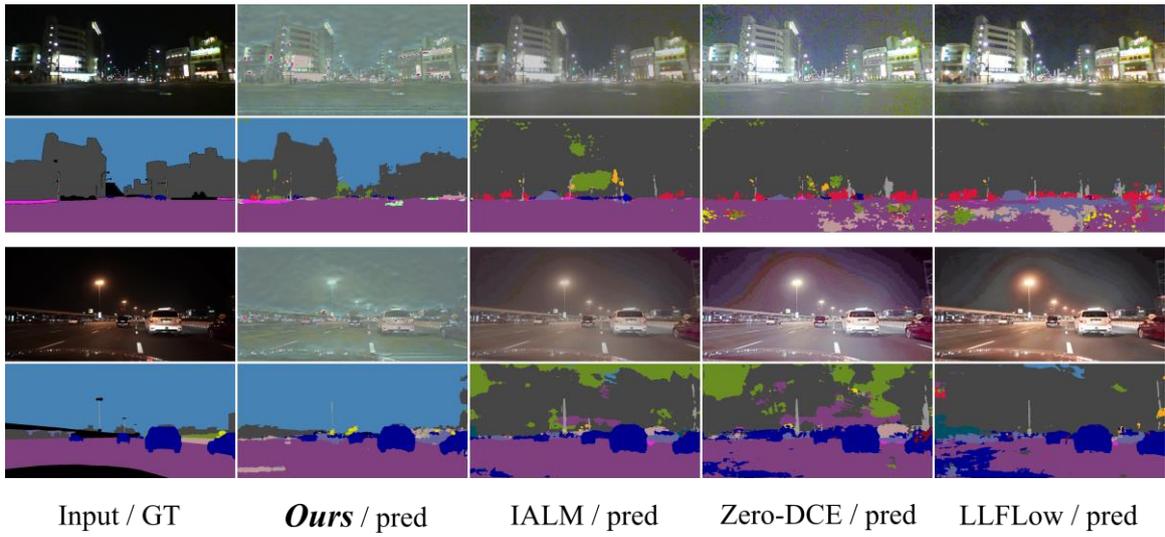

(a), Comparison of enhanced images obtained by each low-light image enhancement method with the predicted results.

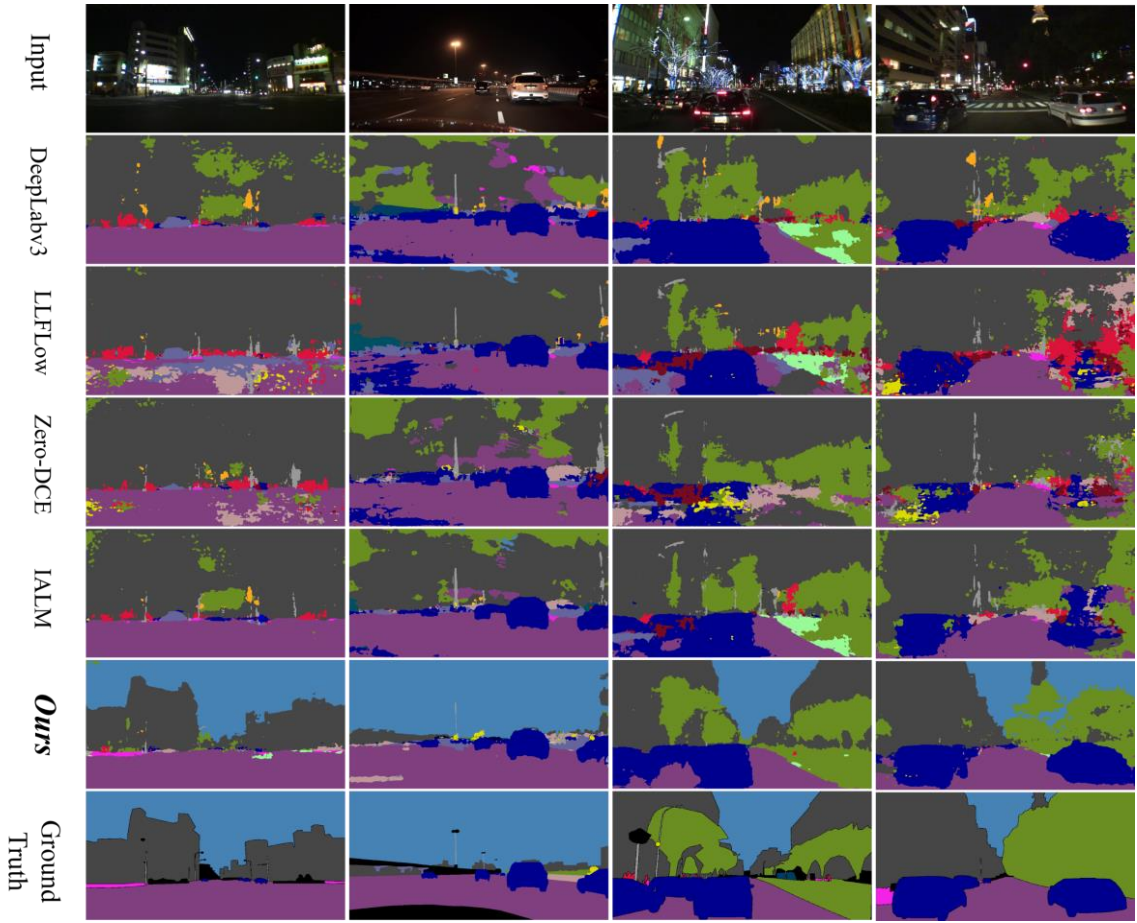

(b), Prediction results for images enhanced by each method.

Figure 3: Qualitative evaluation results of the semantic segmentation task.

If PAM can approximate the image processing performed by GEM, we predicted that the mIoU, when trained with PAM alone, would be consistent with that when trained with both GEM and PAM.

We show an experimental result in Table 5. According to Table 5, training only GEM, while omitting the PAM, yields superior improvement in the performance of the downstream recognition model compared to conventional methods. However, the performance improvement was still marginal, indicating the limitation of the performance improvement with adopting only global adjustment of the input image.

On the other hand, when only the PAM was trained without the GEM, the accuracy is dramatically improved. This demonstrated the advantage of adopting pixelwise image correction.

However, the performance when only PAM was trained did not achieve the performance when both GEM and PAM were trained. This demonstrated the advantage of explicitly splitting and introducing Global image adjustment and Pixelwise Image Adjustment.

In principle, replacing PAM with a deeper fully convolutional neural network could implicitly approximate the processing of GEM in the previous stage. However, by making global image adjustment and pixelwise adjustment independent, a lightweight convolutional neural network alone can effectively improve the performance of the downstream recognition model.

Through the above experiment, we have demonstrated the effectiveness of the two proposed modules.

Table 5: Quantitative results of the ablation study.

| Model | mIoU ↑ |
|---|---|
| DeepLabv3+ | 18.4 |
| Ours (w/o PAM) and DeepLabv3+ | 21.9 |
| Ours (w/o GEM) and DeepLabv3+ | 31.6 |
| ***Ours and DeepLabv3+*** | ***34.4*** |

## 5 DISCUSSIONS

We proposed a novel low-light image enhancement method aimed at improving the performance of downstream recognition models. To evaluate the effectiveness and generalizability of our proposed method, we conducted experiments using two recognition tasks: single person pose estimation and semantic segmentation.

The experimental results indicate that the images corrected by our proposed method may contain artifacts, and thus may not be suitable for human vision. However, they were shown to improve recognition performance compared to other methods. This suggests that high image quality may not necessarily required to improve recognition performance.

Furthermore, comparing the images corrected by the proposed method for the two tasks, it was unclear which factors mainly contributed to the improved performance. When the recognition model by Lee et al. for pose estimation was used, purple artifacts appeared in the output images. In contrast, using DeepLabV3+ as the recognition model resulted in output images with a greenish tint. The images corrected by our method had unique characteristics for each downstream recognition model, but these characteristics were not maintained when the recognition model was changed.

These findings suggest that the images comprehensible for recognition models may not share a universal feature, but images corrected for each recognition model end up having their own set of unique features corresponding to the respective recognition model.

## 6 CONCLUSIONS

We proposed a novel low-light image enhancement method aimed at improving the performance of image recognition models based on neural networks. Our proposed method consists of two modules: the Global Enhance Module (GEM) and the Pixelwise Adjustment Module (PAM). The GEM derives exposure correction parameters from resized low-resolution images to adjust the exposure and the color balance of the input image globally. The PAM takes the exposure-corrected image from the GEM and predicts pixel-level features that are effective for downstream recognition models, thereby enhancing the image effectively. The entire framework is trained end-to-end, optimizing the low-light image enhancement model to minimize model specific loss, enhancing its performance.

Notably, our proposed method is composed entirely of a lightweight convolutional neural network, containing only 577k parameters. This makes it significantly lighter compared to conventional low-light image enhancement methods. As a result, using our method as a front-end filter allows for the improvement of recognition performance under low-light conditions without the need to retrain various existing pretrained recognition models. Experiments across two tasks utilizing low-light images demonstrated that applying our method to recognition models achieved higher performance than conventional methods, confirming its effectiveness and generalizability.